\pdfoutput=1

\documentclass[11pt]{article}

\usepackage{acl}

\usepackage{times}
\usepackage{latexsym}
 \usepackage{graphicx} 

 \usepackage{enumitem}
 \usepackage[utf8]{inputenc}
\usepackage{inconsolata}
\usepackage{booktabs}
\usepackage{multirow}
\usepackage{framed}
\usepackage{tikz}
\usepackage[T1]{fontenc}
\usepackage{microtype}

\usepackage{pdfpages}

\title{Speech vs. Transcript: Does It Matter for Human Annotators in Speech Summarization?}

\author{Roshan Sharma$^{1}$\thanks{Author is now at Google}, Suwon Shon$^{2}$, Mark Lindsey$^{1}$, Hira Dhamyal$^{1}$, Rita Singh$^{1}$ \and Bhiksha Raj$^{1,3}$  \\
  $^{1}$Carnegie Mellon University, USA $^{2}$ASAPP Inc, USA \\
  $^{3}$Mohammed bin Zayed University of AI, Abu Dhabi \\
  \texttt{shroshan@google.com}}

\begin{document}
\maketitle
\begin{abstract}
Reference summaries for abstractive speech summarization require human annotation, which can be performed by listening to an audio recording or by reading textual transcripts of the recording. In this paper, we examine whether summaries based on annotators listening to the recordings differ from those based on annotators reading transcripts. Using existing intrinsic evaluation based on human evaluation, automatic metrics, LLM-based evaluation, and a retrieval-based reference-free method. We find that summaries are indeed different based on the source modality, and that speech-based summaries are more factually consistent and information-selective than transcript-based summaries. Meanwhile, transcript-based summaries are impacted by recognition errors in the source, and expert-written summaries are more informative and reliable.  
We make all the collected data and analysis code public\footnote{https://github.com/cmu-mlsp/interview\_humanssum} to facilitate the reproduction of our work and advance research in this area.


\end{abstract}

\section{Introduction}
\label{sec:intro}

\begin{figure*}[htb]
    \centering
    \includegraphics[scale=0.6]{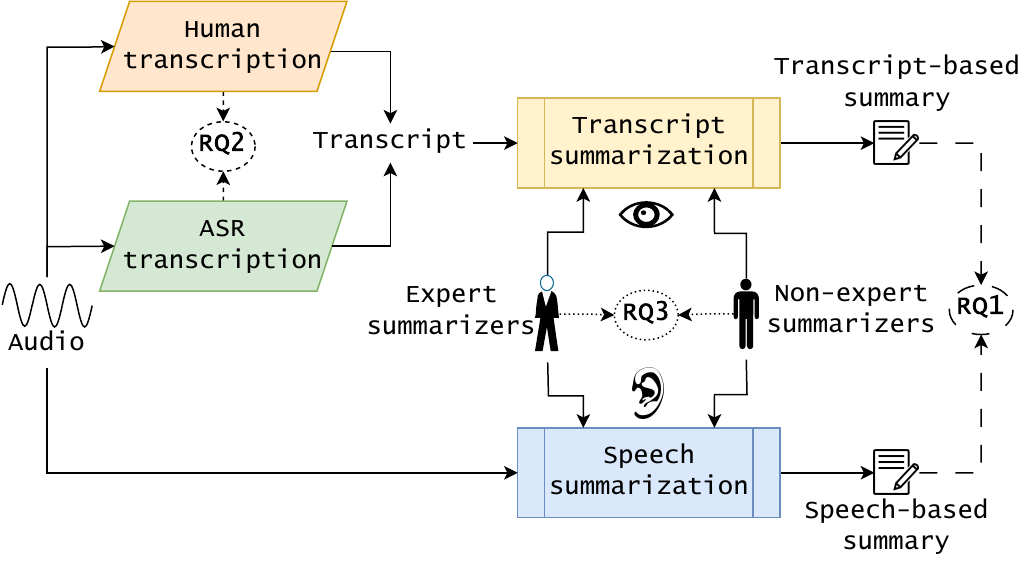}
    \caption{Overview of collection process and research questions}
    \label{fig:overview}
\end{figure*}

Speech summarization ~\cite{hori2002automatic,rezazadegan2020automatic} is the task of analyzing long speech recordings and generating a short textual summary that concisely conveys this information. Abstractive speech summarization ~\cite{sharmaEnd,palaskar21_interspeech} aims to produce summaries that a human would write~\cite{mani1999advances}. Prior work in abstractive speech summarization~\cite{palaskar-etal-2019-multimodal,kano2023icassp,sharma2022xnor,sharmabass} has focused on developing automatic methods.

However, to the best of our knowledge, there has been little work examining how humans perform speech summarization. Such research is particularly important because the goal of abstractive summarization is to produce human-like summaries, and having high-quality human summaries is crucial for learning how to automatically summarize, i.e., for model training and evaluation. Also, information about how humans summarize may facilitate improved automatic approaches, and hence it is important to explore the annotation process. In this paper, we delve deeper into the process of human annotation for abstractive speech summarization.

Fig. \ref{fig:overview} presents an overview of our paper. Humans can summarize speech when it is presented by (a) listening to the speech directly and then constructing the summary based on this, or (b) reading a textual transcript of the content spoken in the recording and using that to construct the summary. It is unclear whether summaries based on listening to audio differ from summaries based on reading transcripts. For transcript-based summarization, the method used to obtain the transcription, either Automatic Speech Recognition (ASR) or manual transcription, may also have a significant effect. Additionally, the quality of a summary, regardless of the source modality, is likely to be determined by the level of expertise of the summarizer. To assess the impact of source modality, transcription, and expertise, we formulate the following research questions. 

\setlist[enumerate]{font={\bfseries}}
\setlist[enumerate,1]{label={(\textbf{RQ}\arabic*)}}

\begin{enumerate}
    \item Are summaries based on annotators listening to speech recordings different from summaries based on annotators reading textual transcripts of speech? If so, how?
    \item If ASR transcripts are used instead of manual transcripts, how do recognition errors impact human summaries?
    \item Do summaries written by non-experts summarizers differ from those written by experts, and if so, do the differences vary based on the source modality?
\end{enumerate}

\setlist[enumerate]{font={}}
\setlist[enumerate,1]{label={(\arabic*}}

Past work in automatic speech summarization suggests that there are acoustic characteristics in speech beyond the textual context that are useful for speech summarization~\cite{kano2023icassp}. This implies that both the transcript and the raw speech carry complementary information that is useful for summarization. The fact that end-to-end models~\cite{sharmaEnd,kano2023icassp,matsuura2023icassp,matsuura2023transfer,sharmabass} that use speech features as input perform differently from cascade models ~\cite{kanoAttention,palaskar21_interspeech,yu-etal-2021-vision,palaskar-etal-2019-multimodal} that use speech transcripts~\cite{ESPnetSUMMSharma2023} suggests that there are differences in automatic summaries resulting from textual or speech input. In this paper, we study whether the same holds for human annotators.

However, any observations on the impact of reading versus listening for annotation are difficult to make from existing corpora like How2~\cite{sanabria2018how2}, SLUE-TED~\cite{shon-etal-2023-slue}, Interview~\cite{ESPnetSUMMSharma2023}, and AMI~\cite{mccowan2005ami}, since it is unclear how human annotations were obtained from speech. Therefore, we select 1002 recordings from the \textit{Interview} corpus test set and create a new evaluation set. We collect two expert human annotations for transcript-based summarization and two for speech-based summarization per recording and assess how the resulting summaries differ.

Comparing two given summaries is challenging. Broadly, we conduct three types of evaluation: (a) comparing the summary to its source transcript (\emph{source-based evaluation}) to establish how extractive and factual the summary is, (b) evaluating the summary independently for its structure and grammar (\emph{structure evaluation}), and (c) by comparing different summaries (\emph{summary comparisons}) to assess relevance and which represents the input better using semantic similarity, LLM, and human evaluation.   Experiments show that speech-based summaries are more information-selective and factually consistent while transcript-based summaries are preferred by human and LLM evaluators. 

In real-world settings, human-annotated manual transcripts are not available, and, therefore, Automatic Speech Recognition (ASR) models may be used to generate such transcriptions. ASR transcripts may contain errors, and we investigate the impact of these errors on transcript-based summarization. To do this, we consider 100 recordings from the Interview test set, obtain speech transcriptions using Whisper~\cite{radford2022robust}, and perform human summarization using 2 expert annotators for each modality.  Experiments show that ASR errors in the transcript degrade the quality of human summaries, resulting in lower coherence, fluency, and factual consistency with the source. 

Prior work in human evaluation of abstractive text summarization highlights the importance of expertise for summary evaluations~\cite{Fabbri_tacl_SUMMEval,gillick-liu-2010-non}. However, to the best of our knowledge, there is no prior work that examines the need for expertise in human speech summarization. Therefore, in this work, we examine whether crowd-sourced non-expert summaries are different from expert summaries. To address this question, we obtain 2 non-expert annotations using Amazon Mechanical Turk (AMT) for each of the 1002 recordings from the \textit{Interview} test set. From these summaries, we analyze (a) differences between expert and non-expert annotations, and (b) whether any differences can be attributed to the approach taken for summary annotation. Analysis shows that non-expert annotations are less fluent and coherent compared to expert summaries but as factually consistent as expert summaries. 

In summary, through this work, we find that transcript-based annotation is valuable if errors are minimal and longer, more informative summaries are desired, while speech-based annotation is desired for higher information selectivity, factual consistency, and resilience to transcription errors.




\section{Related Work}
\label{sec:related}

For this study, we describe related work in three areas: (a) metrics for comparing summaries, (b) other work investigating the impact of source modality on human performance, and (c) investigations of ASR error impact on Spoken Language Understanding. 

\noindent \textbf{Comparing Summaries - Metrics}: Numerous metrics have been proposed to evaluate the quality of summarization \cite{fabbri2021summeval}. Earlier works use metrics like ROUGE-H, ROUGE-L \cite{zhou2006paraeval}, ROUGE-WE \cite{ng2015better}, BLEU \cite{papineni2002bleu}, sentence recall/precision is commonly used for simple sentence based extraction summaries \cite{hirohata2005sentence}. Some studies have found that ROUGE and BLEU-based metrics correlate well with human judgments on summarization. However such scores are not always the best metric for summarization, especially when the input is longer like in meeting summarization and summarization of scientific articles.
There have been scores of other metrics which are either variants or combinations of the earlier mentioned metrics like BertScore \cite{zhang2019bertscore}, Sentence Mover’s Similarity (SMS) \cite{clark2019sentence}, SummaQA \cite{scialom2019answers}, SUPERT \cite{gao2020supert}, CHRF \cite{popovic2015chrf}, METEOR \cite{banerjeemeteor}, CIDEr \cite{vedantam2015cider}, or model generated ones like s3 \cite{peyrard2017learning}, BLANC \cite{vasilyev2020fill}, summACCY \cite{hori2003evaluation}, which are trained to predict the quality of an input summary. 

LLMs have been used as an alternative method for summary evaluation. While some work expresses doubt that LLMs have reached a human level of summary evaluation capabilities \cite{llm2}, other methods have yielded more optimistic results \cite{llm1}. LLM evaluation methods usually take the form of automatic replacements for human evaluations such as Likert scale scoring, pairwise comparison, and binary factuality comparison \cite{llm4}. LLMs can also be used to answer qualitative questions about the factual consistency of a summary and explain the reasoning behind the provided answers \cite{llm3}.

\noindent \textbf{Source Modality - Transcript vs. Speech:} We are also interested in studying whether input modalities have an impact on task performance. Cognitive psychology literature has studied this question under various settings. \cite{khan2022type} studies cognitive performance and information retention, when note-taking is performed using either voice or text. The paper finds that when the notes are taken using voice versus when they are written down by hand, voice leads to a higher conceptual understanding of the topic. \cite{stollnberger2013input} studies how input modality effects learning of the English language, specifically whether learning from lecture audio vs reading lecture notes impacts retention.
Other papers examine the interaction between input modality (text vs speech) and what users prefer when performing a task. Studies have reached various conclusions regarding the most useful/important modality based on the task at hand, what we do find is that the difference exists and is significant. 

\noindent \textbf{Impact of ASR on Downstream Tasks:} ASR errors are known to have an impact on downstream task performance, like emotion recognition \cite{schuller2009emotion,feng2020end}, spoken language understanding~\cite{shon-etal-2023-slue}, and Natural language understanding and dialog management \cite{serdyuk2018towards, shivakumar2019spoken}. Therefore, it is important to assess their impact on human summarization.

\section{Data Collection}
\label{sec:data}
Datasets that are used for speech summarization today do not have information on which of the two approaches, i.e., reading a textual transcript of the recording or listening to the recording to summarize was employed to obtain annotation. Further, they only have one annotation per recording, which may inhibit analysis due to annotator-specific biases or tendencies. Therefore, to conduct our experiments, we curate a multi-reference dataset from an existing dataset. 

We considered a subset from the test partition of the \emph{Interview}~\cite{ESPnetSUMMSharma2023,majumder-etal-2020-interview,zhu-etal-2021-mediasum} corpus for speech summarization. \textit{Interview} is the largest open-domain corpus with around 5-minute-long speech recordings containing spontaneous speech along with real audio noises representing music, and events inside and outside the studio corresponding to discussion topics\footnote{The authors acquired permission from NPR to use NPR data for this research}. Speech recordings contain spontaneous speech along with real audio noises that represent music, and events inside and outside the studio corresponding to discussion topics. The dataset has 49.4k recordings that span multiple topics and domains, with an average recording length of 5.28 minutes. The transcription on average contains 886.4 words and the summary contains 40.1 words on average, representing a compression ratio of 22.09 between the transcript and summary. Around 48\% of words in the summary are not present in the transcription, making this a relatively challenging task. On average, each recording has 24 turns and 4 speakers. 


Ensuring high-quality summary annotations is crucial to making reliable observations. Therefore, we took the necessary steps to ensure a fair and reliable data collection pipeline. Initially, we consulted with an expert from the in-house data collection team and conducted small pilot studies to make design decisions.

\noindent
\textbf{Audio Length}: We conducted in-house pilot studies comprising six questions by trying to summarize audio recordings that were 2 minutes and 5 minutes long. As expected, summarizing 5-minute recordings consumed more time and effort, and was harder due to the high compression ratio. We concluded that limiting the audio length to 2 minutes would provide sufficient content to summarize without increasing the difficulty level of summarization significantly. Using forced alignment, we obtained audio reference transcripts up to the first 2 minutes. 

\noindent
\textbf{Summary Length}:
A hard limit was not set on the length of the summary, but it was suggested that summaries should contain at least 2 sentences to convey the essence of the conversation. It was recommended that summaries contain between 50 to 80 words for the 2-minute long recordings.

\noindent \textbf{Data Selection}:
To obtain reasonable summaries, it is important that audio recordings and textual transcripts are understandable, and have a sufficient amount of information to summarize. In cases where background noise or music is present, or when a large portion of the recording does not contain speech, these criteria are violated, leading to sub-optimal recordings for summarization. Therefore, audio recordings were transcribed using \texttt{Whisper-medium} ~\cite{radford2022robust}, and the resulting Word Error Rate (WER) was used to remove unsuitable audio recordings from consideration. Based on manual inspection, a threshold of 25\% WER was set, which eliminated recordings with only music or large amounts of noise.

\noindent \textbf{Expert and Non-expert}: Our data comprises 1,002 recordings in total, with each recording having a speech recording and its corresponding reference transcript. For each recording, we collected two expert summaries and two non-expert summaries by reading the text or listening to the audio, which gave us a total of 4 summaries based on the text and 4 summaries based on the audio on the same recording.  Additionally, we solicited summary annotations from speech recognition transcripts for 100 of these 1002 recordings, sampled at random so the Word Error Rate with \texttt{Whisper-medium} is non-zero, i.e., the transcript has ASR errors.

\noindent
\textbf{Annotator Overlap}:
It is worth noting that each annotator was assigned to work on either the transcript or audio recording for a particular recording, but not both. This approach ensured that no annotator saw audio or transcription for a recording they had already annotated.

\noindent \textbf{Expert Summary Annotation}: Expert summary annotations were obtained through a third-party vendor that passed in-house qualitative assessments of summary outputs from a pilot annotation. The third-party vendor had proven their quality of annotation by consistently delivering similar summary annotations for other projects of the team. Detailed annotation guidelines are in Appendix \ref{sec:appendix_expert}.

\noindent \textbf{Non-Expert Summary Annotation} Non-expert summary annotations were obtained through the Amazon Mechanical Turk (MTurk) platform. Each HIT required the annotator to accept the terms of a consent form that explained the task and requested consent to share responses for reproducible research. Each HIT included four questions, where annotators were asked to write 2 summaries by listening to audio recordings, and two summaries by reading transcript passages. More details are provided in Appendix ~\ref{sec:appendix_nonexpert}.

\section{Evaluating Differences across Summaries}
\label{sec:evaluation}
Summary evaluation is an inherently challenging problem for general settings since there is often no clear definition of what constitutes a good summary. Within this paper, we deem a summary good if it is factually faithful to and representative of the input source while being abstractive, coherent, and fluent.  Therefore, in this section, we outline three types of evaluations. 

\subsection{Source-based Evaluation}
\noindent \textbf{Summary Length and Compression Ratio}: Summary length is measured in words and the compression ratio \cite{mani1999advances} is defined as the number of words in the reference transcript divided by the number of words in the summary. The greater the summary length, typically, the higher the amount of information contained in the summary, and the lower the compression ratio.

\noindent \textbf{Novel Words (\%)}: The percentage of words in the summary that are not present within the source transcript, which could be considered to measure the extent of paraphrasing within the summary. 

\noindent \textbf{Extractiveness}: It is a measure of how extractive the summary is and can be approximated by computing lexical overlap using ROUGE-L~\cite{rouge} between the source transcript and the summary. 

\noindent \textbf{Informativeness-Entities}: We propose to compare named entities predicted from the source text and summaries as a proxy for the amount of information contained within the summary. Named entities are extracted using the Entity Recognizer\footnote{https://spacy.io/api/entityrecognizer} from Spacy~\cite{spacy2}.

\noindent \textbf{Informativeness-Semantic Similarity}: Also, we compare semantic similarity between the source and summaries using metrics like BERTScore~\cite{zhang2019bertscore} and BARTScore~\cite{yuan2021bartscore}. 

\noindent \textbf{Retrieval-based informativeness:} A summary can be considered representative of or informative about the source when it can discriminate the correct source from a selection of all sources. We describe a retrieval-based measure where we compute the semantic embedding similarity (using BERTScore) between the hypothesis summary and all possible source texts. A discriminative summary should produce the highest similarity between the summary and the \emph{correct} source text. This is measured using retrieval accuracy (RAcc). We also report the Mean Reciprocal Rank (MRR) as an indicator of retrieval performance.

\noindent \textbf{Factual Consistency:} We use UniEval to estimate the factual consistency of summaries with respect to the source transcript. 

\subsection{Structure Evaluation}
To evaluate coherence and fluency automatically, we leverage UniEval~\cite{zhong-etal-2022-towards}.

\subsection{Summary Comparisons}
\begin{figure}
    \centering
    \includegraphics[scale=0.4]{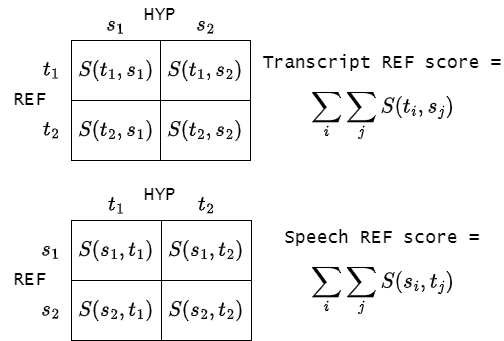}
    \caption{Pairwise Score Computation}
    \label{fig:pairs}
\end{figure}

\noindent \textbf{Pairwise Similarity}: We compute relevance and consistency using UniEval~\cite{zhong-etal-2022-towards}, and use BARTScore as a measure of semantic similarity.  Pairwise scoring is illustrated in Fig. \ref{fig:pairs}, where $s_i$ is a speech-based summary, $t_j$ is a transcript-based summary, and $S(a, b )$ calculates the score between reference $a$ and hypothesis $b$ (e.g., BART score). The idea behind pairwise scoring is to compare two types of summaries, where one type is a reference and the other a hypothesis. This process results in two scores, a speech reference score, and a transcript reference score. By comparing these two scores, we can identify cases when the information in one type of summary is present within the other type of summary.

\noindent \textbf{Pairwise Inter Annotator Agreement (IAA)}: For text generation tasks, the inter-annotator agreement is challenging to quantify. We use a combination of lexical and semantic similarity metrics like ROUGE-L~\cite{rouge}, and BARTScore~\cite{yuan2021bartscore} to automatically represent inter-annotator agreement. 

\noindent \textbf{Factual Consistency} and \textbf{Source Representation}: We employ LLM-based evaluation and human evaluation to estimate the factual consistency of each summary, and how well each summary represents the source. These evaluation methods are described in detail below.

\label{sec:appendix_llmeval}
\begin{figure*}[h]
\begin{tikzpicture}[every node/.style={draw,text width=0.95\textwidth,minimum width=0.95\textwidth}]
\node {
\scriptsize
\texttt{The following are two summaries of the given source text:}

\texttt{Summary 1: [Summary 1]}

\texttt{Summary 2: [Summary 2]}

\texttt{Source: [Source]}

\centering \underline{Factual Consistency Question}

\raggedright
\texttt{Answer the following multiple-choice question:} \\
\texttt{What is the most appropriate statement about the summaries?} \\
\texttt{(a) Summary 1 is more factually consistent with the source text than Summary 2.} \\
\texttt{(b) Summary 2 is more factually consistent with the source text than Summary 1.} \\
\texttt{(c) Both Summary 1 and Summary 2 are equally factually consistent with the source text.} \\
\texttt{Give your answer as ``Answer: a'', ``Answer: b'', or ``Answer: c''. Then, start a new line and explain your reasoning in a single paragraph following your answer.} \\

\centering \underline{Source Representation Question}

\raggedright
\texttt{Answer the following multiple choice question:} \\
\texttt{What is the most appropriate statement about the summaries?} \\
\texttt{(a) Summary 1 is more representative of the source text than Summary 2.} \\
\texttt{(b) Summary 2 is more representative of the source text than Summary 1.} \\
\texttt{(c) Both Summary 1 and Summary 2 are equally representative of the source text.} \\
\texttt{Give your answer as ``Answer: a'', ``Answer: b'', or ``Answer: c''. Then, start a new line and explain your reasoning in a single paragraph following your answer.} \\
};
\end{tikzpicture}
\caption{\centering Template for questions presented to ChatGPT. Similar questions are used to solicit human scores.}
\label{fig:fact_question}
\end{figure*}

\subsubsection{LLM-based Evaluation}
Following the LLM summary evaluation methods listed in Section \ref{sec:related}, we use ChatGPT (\texttt{gpt-3.5-turbo-0125}) as a qualitative evaluation method for the summaries collected for this work. Specifically, we designed a multiple choice-style AB test to compare summaries in terms of either factual consistency or accurate representation of the source transcript. The two questions posed to the LLM are displayed in Fig. \ref{fig:fact_question}, where the text in the brackets represents the summaries being evaluated and the associated source text. Each question was presented in a separate session to avoid bias from previous questions.

Each question is asked twice for each AB pairing, but with the order in which the summaries are presented reversed to avoid biasing the results. For example, if we were comparing a transcript-based summary and a speech-based summary, the question would be asked one time with the transcript summary as Summary 1, and then another time with the speech summary as Summary 1. Additionally, since all summaries are collected in pairs, we compare each summary to both summaries from the other group, exhausting all possible combinations. For example, if we were comparing a pair of transcript-based summaries to a pair of speech-based summaries, the first transcript summary would be compared to both speech summaries, and then the second would be compared to both speech summaries, resulting in four total comparisons. Considering both the order of presentation and the comparison combinations, eight questions are posed to the LLM per source sample.

\subsubsection{Human Evaluation}
To validate the expert summary comparison, we also conducted an AB test for 30 recordings with 15 annotators. We present 2 summaries based on different modalities and their source transcript to a user. We designed 2 multiple-choice questions that ask human annotators to compare summaries based on their factual consistency and how accurately they represent the source transcript.
The setup of questions is similar to that in LLM evaluation.

\section{Research Methodology}
\label{sec:method}
\subsection{RQ1: Speech versus Text Inputs}
We apply the methods and metrics described in Section \ref{sec:evaluation} to explore differences between summaries derived from speech recordings and summaries extracted from text transcripts. We use the expert annotations obtained for 1002 recordings for this analysis since expert summaries are more reliable.

\subsection{RQ2: Impact of Transcript Errors on Human Summarization}
We measure the impact of ASR error propagation on the output summary by comparing a subset of 96 expert summaries given ASR-generated transcripts and ground-truth transcripts. The summaries resulting from the Whisper-generated transcripts are compared to the summaries based on ground-truth transcripts using the methods and metrics outlined in Section \ref{sec:evaluation}. 

We analyze the effect of ASR errors on the quality of the summaries based on ASR transcripts by plotting informativeness measures like Entity-F1 and WER, and find that as WER increases, Entity-F1 decreases.

\subsection{RQ3: Expert versus Non-Expert Annotation}
We use all methods and metrics from Section \ref{sec:evaluation} to explore the effect of expertise on summary quality. Similar to RQ1, we provide an overall comparison across speech-based summaries as opposed to transcript-based summaries arising from non-expert annotations and conclude on the reliability of expert and non-expert summaries.

\section{Experimental Results and Discussion}
\label{sec:results}

\begin{table*}[h]
    \small
    \centering
    \begin{tabular}{lcc|cc}
    \toprule
          & \multicolumn{2}{c}{Expert} & \multicolumn{2}{c}{Non-expert} \\ 
          Metric & Transcript & Speech & Transcript & Speech \\
          \toprule 
          Summary Length &  78.58 $\pm$ 27.72&  56.95 $\pm$ 18.30& 75.35 $\pm$	17.70	& 77.42 $\pm$ 33.85\\
          Compression Ratio &3.76 $\pm$ 2.20& 5.20 $\pm$ 3.06 & 3.69	$\pm$	1.95	& 3.66$\pm$		1.86	\\
          Word Vocabulary Size &20,408&  16,211& 21511	&	21676\\
          Novel Words \%&24.63	$\pm$ 13.57&  25.66 $\pm$ 13.22 & 37.47	$\pm$ 13.38&	38.45$\pm$	14.21\\ 
          \midrule
          Extractiveness$\downarrow$ & 20.70 $\pm$ 10.53&  \textbf{18.57	$\pm$ 10.88}&\textbf{18.37$\pm$	7.26}&	18.38	$\pm$11.29\\
          Entity F1$\uparrow$ &\textbf{33.63	$\pm$ 29.83}&  26.21	$\pm$ 26.90& \textbf{34.51	$\pm$ 17.28}&	30.09$\pm$	17.72\\
          BERTScore $\uparrow$ & \textbf{85.66	$\pm$2.45} &  85.32	$\pm$2.42& \textbf{85.28$\pm$	2.32} &	85.06 $\pm$2.76\\
          BARTScore $\uparrow$ & \textbf{-3.37	$\pm$0.47 } &  -3.46 $\pm$ 0.45& \textbf{-3.41$\pm$	0.44}&	-3.43$\pm$	0.53\\
          Retrieval Accuracy $\uparrow$ & \textbf{68.86 + $\pm$ 46.31}	&	59.04 $\pm$ 49.75 & \textbf{71.82 $\pm$ 44.99} &  65.78 $\pm$ 47.44  \\
          MRR $\uparrow$ & 70.23 $\pm$ 16.08 & \textbf{75.74 $\pm$ 16.29} & \textbf{76.56 $\pm$ 71.01} & 71.01 $\pm$ 41.57 \\
          Factual Consistency (UniEval) $\uparrow$ & 82.68	$\pm$ 16.38&	\textbf{84.58$\pm$	15.72}&	\textbf{82.08	$\pm$ 16.90}	& 78.57$\pm$	19.79\\ 
          \midrule
          Fluency (UniEval) $\uparrow$ & \textbf{94.02$\pm$	4.54}	&94.01$\pm$	5.26&	\textbf{92.90$\pm$	8.21} & 92.65	$\pm$9.35\\
          Coherence (UniEval)$\uparrow$ &\textbf{91.82$\pm$	12.97}&	89.34$\pm$	15.63	&\textbf{90.88$\pm$	15.19}&	87.95$\pm$	19.57\\ \midrule 
          Pairwise Rep. (LLM) $\uparrow$ & \textbf{66.67} & 28.52 & - & -\\
          Pairwise Factualness (LLM) $\uparrow$ & \textbf{60.16} & 33.07 & - & - \\ 
          Pairwise Rep. (Human) $\uparrow$ & \textbf{48.57} & 21.43 &- & -\\ 
          Pairwise Factualness (Human) $\uparrow$ & \textbf{39.52} & 22.62 &- & -\\
          IAA - ROUGE-L $\uparrow$ & 24.92 $\pm$ 7.82 & \textbf{28.98 $\pm$ 9.29} & \textbf{32.67 $\pm$ 12.56} & 28.62 $\pm$ 12.52 \\
          IAA-BARTScore $\uparrow$ & \textbf{-2.71 $\pm$ 0.53} & -2.55 $\pm$ 0.55 & \textbf{-2.72 $\pm$ 0.61} & -2.93 $\pm$ 0.73 \\
          
          Pairwise BARTScore $\uparrow$ & -2.81$\pm$0.51&\textbf{-2.56$\pm$	0.52}&	\textbf{-2.85$\pm$	0.63 }&	-2.88$\pm$	0.70\\ 
          \bottomrule
    \end{tabular}
    \caption{RQ1 and RQ3 Evaluation: Speech-based versus Transcript-based Summaries for Expert and Non-expert annotators}
    \label{tab:expert_rq13}
\end{table*}

\subsection{RQ1: Speech versus Transcript Inputs}

Evaluation across the expert summaries, including a comparison between transcript-based and speech-based summaries, is shown in the first two columns of Table \ref{tab:expert_rq13}.

It is immediately apparent that speech-based summaries are significantly shorter and more compressed than transcript-based summaries, indicating that speech-based summaries are highly selective. This is likely because the summarizers extract only the main points when listening to a recording, while a transcript serves as an easier reference for smaller details. This claim is supported by the low level of extractiveness in the speech-based summaries, indicated by a higher percentage of novel words, lower ROUGE-L scores, and a lower percentage of retained named entities. Additionally, coherence for speech-based summaries is relatively low, likely because they are not long or detailed enough to form a very structured summary.

These results also indicate that speech-based summaries are less informative than transcript-based summaries. Again, this is likely because summarizers can extract details by referencing transcripts much more easily than by searching through the recording. Results that support this claim are high BERT- and BARTScores, high retrieval accuracy, factual consistency, and human and LLM ratings. One interesting observation from the results is that the source retrieval task based on semantic similarity has higher standard deviation scores compared to those observed for other metrics. We note that the proposed retrieval task is challenging compared to simply obtaining similarity scores. Of the source texts in the dataset, some of them are more closely related than others. Hence some retrieval problems are much harder than those that involve retrieval among dissimilar or less similar sources. This phenomenon leads to a higher standard deviation for our retrieval metrics. 

The higher pairwise BARTScore for speech references indicates that more of the information within the speech summary is likely present in the transcript-based summary, making speech-based summaries more selective in relevant information. IAA scores computed based on ROUGE-L and BARTScore indicate that speech-based summaries promote higher consensus between raters. 

The only result that did not exhibit statistically significant influence from the source modality was fluency. This is reasonable, as the fluency of the model is determined more by the language proficiency of the individual writing the summary than the content.

For \textbf{RQ1}, we find that \emph{speech-based summaries seem to be more selective, factually consistent, and perhaps more abstractive than transcript-based summaries, while transcript-based summaries contain more information extracted directly from the source than their speech-based counterparts}.

\subsection{RQ2: Impact of Transcript Errors on Human Summarization}

\begin{table*}
    \small
    \centering
    \begin{tabular}{lcccc}
    \toprule
            Metric& Transcript Overall&  Speech Overall & ASR Overall  \\\toprule 
            Summary Length & \textbf{85.29$\pm$40.58}	 &56.94$\pm$36.89 &	73.67$\pm$36.89  \\
            Compression Ratio & 0.40$\pm$	0.73&	\textbf{0.52$\pm$	0.81}	& 0.27$\pm$	0.85\\
             \midrule
            Extractiveness (ROUGE-L with Transcript)& 21.94$\pm$10.95	& \textbf{19.17$\pm$11.60}	&19.25$\pm$9.33 \\
            Entity F1& \textbf{35.17	$\pm$18.93 }&26.28 $\pm$ 15.65	&30.24$\pm$17.67  \\
            BERTScore with Transcript& \textbf{85.92	$\pm$ 2.55}&85.56$\pm$2.55&	85.49$\pm$2.45& \\
            BARTScore with Transcript&\textbf{-3.33 $\pm$ 0.48}&	-3.44$\pm$0.47	&-3.41 $\pm$0.45\\
            Retrieval Accuracy&\textbf{85.42$\pm$	35.29}&	71.88$\pm$	44.96	&77.60$\pm$	41.68 \\
            MRR& \textbf{89.09$\pm$	27.55}&	79.44	$\pm$34.27&	83.49$\pm$32.11 \\
            Factual Consistency (UniEval)& 83.56 $\pm$ 14.45 &	\textbf{86.16$\pm$13.83}&	83.48$\pm$14.25\\
            \midrule
            Fluency (UniEval)& 94.31 $\pm$ 3.62 &	94.23 $\pm$ 3.90 &	\textbf{94.36 $\pm$ 3.72}\\
            Coherence (UniEval)&\textbf{91.51 $\pm$ 14.90}&	88.64 $\pm$ 16.04	&	90.81 $\pm$ 13.34\\ \midrule
            Pairwise Rep. (LLM)& \textbf{40.06} & 21.57 & 33.81\\
            Pairwise Factualness (LLM)& \textbf{37.76} & 22.74 & 32.81 \\ 
            Pairwise BARTScore&-2.81$\pm$ 0.48 & \textbf{-2.59 $\pm$ 	0.51} &	-2.79 $\pm$ 0.51 \\ 
            \bottomrule
    \end{tabular}
    \caption{RQ2 Evaluation: Impact of Transcript Errors on Human Summarization}
    \label{tab:asr_results}
\end{table*}

\begin{figure}[h]
    \centering
    \includegraphics[scale=0.5]{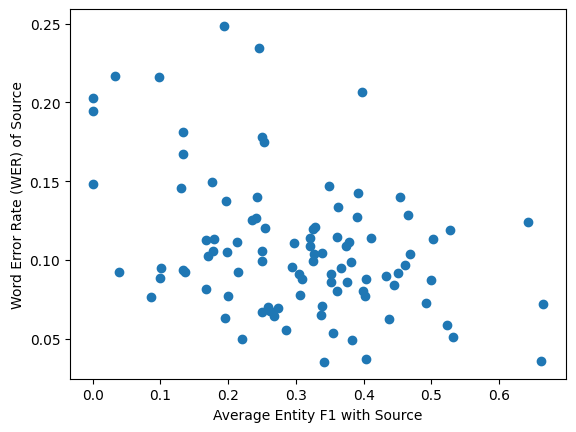}
    \caption{Scatterplot showing variance of entity F1 of summary compared to source reference transcript with Word Error Rate (WER) of audio transcription}
    \label{fig:wer_entity}
\end{figure}

Table \ref{tab:asr_results} compares the summaries based on the subset of 100 recordings. It is interesting to compare the summaries based on ASR transcriptions to the summaries based on manual transcriptions. Nearly all indicators of informativeness are significantly higher for the manual transcriptions than the ASR transcriptions. This is to be expected, as propagated recognition errors can directly affect measures like Entity F1, factual consistency, and semantic similarity. Coherence is similar for the two types of transcripts, which is consistent with the idea that the ability to reference text allows the summarizer to capture the details and structure of the source.

Comparing ASR-transcript-based summaries to speech-based summaries, we see similar trends to those in RQ1. Speech-based summaries are still shorter, more selective, less informative, more factually consistent, and likely more abstractive than ASR-transcript-based summaries. In general, LLM evaluation indicates that ChatGPT prefers transcript-based summaries over speech-based summaries, even if the source transcripts contain errors. 

Figure \ref{fig:wer_entity} contains a scatterplot between the Word Error Rate (WER) of the automatic source transcript and average entity F1 of the human-annotated summary compared to this automatic source transcript. A weak negative correlation is observed between WER of the source transcript and Entity F1, showing that as WER increases, Entity F1 scores tend to decrease on average. 

For \textbf{RQ2}, we find that \emph{transcription errors decrease the informativeness and coherence of transcript-based summaries, but do not significantly impact fluency}.

\subsection{RQ3: Expert Annotation versus Non-Expert Annotation}

From Table \ref{tab:expert_rq1}, we observe that the speech-based summary length from expert annotation is significantly lower than its non-expert counterpart, while the expert speech-based and transcript-based summary lengths are very different. The word vocabulary size and the percentage of novel words are lower in the expert summaries (both speech and transcript source) compared to the non-expert summaries. 

Extractiveness is lower for non-expert speech-based and transcript-based summaries compared to the expert ones, referring to the fact that the non-experts do not tend to use phrases directly from the source whereas the experts do. Entity F1 score is higher for non-expert speech-based and transcript-based summaries, which is because non-expert summaries are on average longer than the expert ones and capture more of the entities in the source. BERTscore, BARTscore, retrieval accuracy, and MRR are higher for the expert summaries than the non-expert summaries (for both speech and transcript sources) referring to the fact that non-expert summaries are less informative than the expert summaries. Factual consistency is lower for non-expert summaries pointing to the fact that expert summaries are more factually consistent. Fluency, coherence, and pairwise BARTscores are lower for non-expert summaries demonstrating the reliability of expert summaries. 

We ran t-tests to conclude whether the non-expert speech-based and transcript-based summaries belong to the same distribution and observed based on BARTScore and other metrics that the p-value was high (see Table \ref{tab:non_expert_annotations} in Appendix \ref{sec:appendix_res_nonexp}). This potentially indicates that non-expert annotations based on speech and transcripts may be very similar and that some annotators may have used similar methods or tools for the two types of annotations. Based on a human examination, it appears that some annotators did use generative AI to respond to annotation requests through Amazon Mechanical Turk, however, such conclusions may not be definitive since such models can generate human-like speech. This makes non-expert summaries less reliable for analysis of speech-based versus transcript-based summaries, and hence, we decided against running the LLM and human evaluation for these non-expert summaries. 

For \textbf{RQ3}, we find that \emph{expert annotations are more informative, coherent, fluent, and less information variant across speech-based and transcript-based summaries}.

\section{Conclusion}
\label{sec:conclusion}
In this paper, we take the first steps towards understanding human annotation for speech summarization by proposing and assessing different approaches. Annotators can either read a transcript of spoken content or listen to the audio recording to produce an abstractive speech summary. We formulate three research questions to assess differences in summaries resulting from variations in source modality (listening to audio versus reading a transcript of spoken content), the nature of transcription (manual versus ASR-based), and the expertise of annotators. Extensive analysis using automatic and human evaluation reveals that speech-based summaries are more information-selective, factually consistent, and abstractive compared to transcript-based summaries. Analysis of the impact of ASR on transcript-based summarization shows that errors in transcript generation decrease the informativeness and coherence of transcript-based summaries. Finally, analysis of expert and non-expert annotations demonstrates higher informativeness, coherence, and fluency across expert annotations. 

Our experiments with non-expert human annotation of transcripts showed that using platforms like Amazon Mechanical Turk may result in less reliable human annotations owing to annotators' tendency to use generative models like ChatGPT for annotations. Further, our work reaffirms that errors in speech transcription not only impact automatic approaches for summarization but also human annotation. This demonstrates the need for reliable transcripts for speech summarization annotation.

Through this paper, we also develop and release a new human-annotated dataset to enable reproduction and encourage further research in the area. We hope that this study galvanizes further research into human summarization, and powers informed model design for automatic speech summarization.

\newpage 

\section*{Ethics Statement}
We have made a concerted effort to make a positive ethical impact during this work. In data collection, we received permission through the proper channels (IRB) and treated all participants as fairly and respectfully as we could. Through the annotation process, we have valued diversity of opinion and annotator backgrounds and compensated any workers at wages beyond the minimum wage in the US. We have also tried to remove elements of bias by using automatic and manual verifications as appropriate. In the technical work, we have been honest and transparent. We have also carefully considered the implications and potential applications of our work and findings, and we do not believe they present any significant risks to individuals or society as a whole.

\section*{Limitations}

A challenge in the context of this work is that existing methods are designed to evaluate summaries of written text, and may have a bias against speech-based summaries. While some of the methods presented here may begin to address this bias (e.g., retrieval accuracy and IAA metrics), the problem persists. Text-based measures effectively evaluate lexical content, but do not measure properties like emotion, prosody and vocal emphasis which may carry information that is useful for speech summarization. Developing a metric that can be used for both speech- and text inputs remains an ongoing research problem with no consensus solution at this time. Therefore, we utilized the available and broadly accepted text-based evaluation metrics in this paper. Despite the potential bias against speech input based summaries, we still find that in certain aspects, i.e., information selectivity, factualness and abstractiveness, speech-based summaries are beneficial over transcript-based summaries. 

Another potentially significant limitation of this paper is the results regarding non-expert summaries. As stated previously, it is possible that the participants on Amazon Mechanical Turk used generative AI software to produce the summaries, so the non-expert analysis is probably not representative of real non-expert summaries.

\section*{Broader Impact}
Our work represents the first of hopefully many forays into understanding human annotation for speech summarization. The notion of understanding how humans perform the task, and what would enable them to produce the best possible annotations is both important and understudied. We hope that our new dataset and the analysis we have presented in this work serve to ignite discussions around community standards for speech summarization annotation. Further, we hope that the strengths and weaknesses of summaries based on different approaches inform developers on how to design and deploy the next generation of speech summarizers and more generally speech foundation models.

\bibliography{anthology,custom}

\appendix

\newpage
\onecolumn
\section*{Appendix}
\label{sec:appendix}
\section{Expert annotation details}
\label{sec:appendix_expert}

\subsection{Annotation guidelines}

Figure \ref{fig:annotation_guideline} shows the detailed guideline we provided for expert annotation to third-party vendors.

\begin{figure*}[h]
\begin{tikzpicture}[every node/.style={draw,text width=0.95\textwidth,minimum width=0.95\textwidth}]
\node {
\scriptsize
The summaries generated for this task have fewer limitations on style and content. Generally speaking, the summaries ought to capture the main topic of the interaction, but we’re otherwise not too caught up on prescribing a particular way of writing, such as maintaining a particular tense throughout.

Annotator selection
- A single annotator should only interact with each segment once, either as an audio file or as a text transcript.

Summary length

- There is no hard limit on maximum summary length, but we expect at least two sentences per summary, minimum and expect 50 to 80 words.
  
Summary content

- The summary should convey the main topic or overarching message contained in the interaction.

- There is nothing that should never be included, but we prefer for annotators to summarize the overall message rather than attempting to incorporate direct quotes from the interaction.

- Do not worry about doing outside research to verify content. It is preferred that the summaries are written with few prior constraints and without overthinking (summaries should be fairly spontaneous).

Summary style

- There is no strict template to follow for these summaries.

- Summaries should maintain a neutral tone.

- Summaries can be as simple or complex as needed to capture the overall message of the interaction.

- While we expect the summaries to be written in complete sentences with proper grammar, no particular tense needs to be maintained.

- Names and places mentioned in the conversations (including the names of conversation participants) can be used, and accurate spellings for those do not matter.

};
\end{tikzpicture}
\caption{\centering Annotation guideline}
\label{fig:annotation_guideline}
\end{figure*}

\subsection{Annotator Compensation and Statistics}

\begin{figure}[h]
    \centering
    \includegraphics[width=0.5\columnwidth]{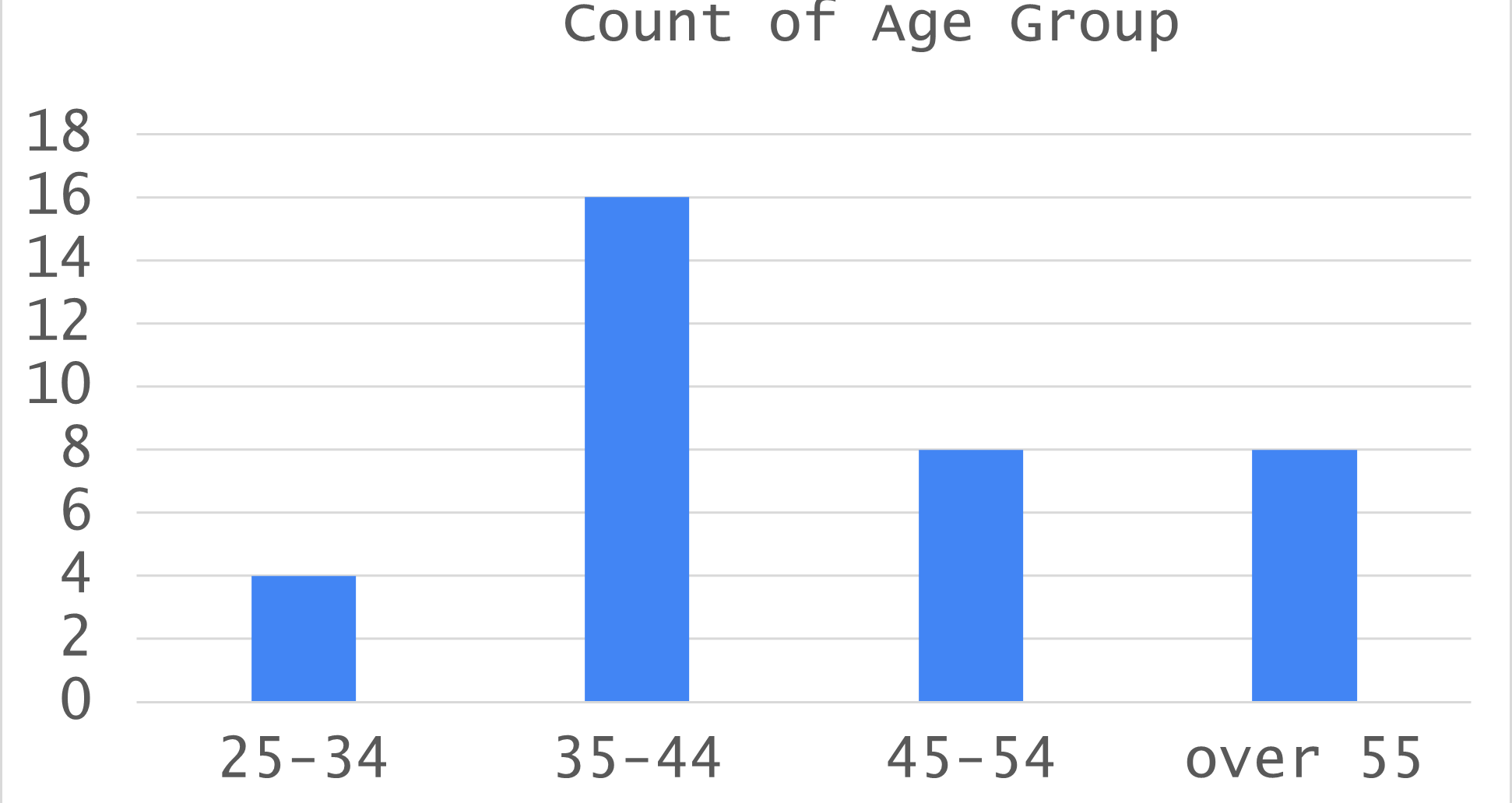}
    \caption{Age distribution for the Expert Annotators}
    \label{fig:age_frequency_expert}
\end{figure}

In total, there are 36 expert annotators. The expert annotators are all American national and native English speakers, proficient in the language, therefore they don't have any difficulty performing the summarization task. Among the 36, 34 are female, one identifies as non-binary and one prefers not to say. There exists some variation in their age group, which is shown in the figure~\ref{fig:age_frequency_expert}.

The third-party vendor passed in-house qualitative assessments of summary outputs from a pilot annotation and contracted to the company's primary third-party vendor for summary annotation. The statement of work has agreements between the company and the vendor, and the company paid 9,341.76 US dollars for a total of 4,208 expert summary annotations. This is a competitive compensation for such expertise, and well beyond the minimum wage within the United States of America, where the annotation was carried out. Annotators consented to the process through the work agreement and were aware of the plans to release the annotations to the public.  

\newpage

\subsection{Example Source and Associated Summaries}

\begin{figure*}[h]
\begin{tikzpicture}[every node/.style={draw,text width=0.95\textwidth,minimum width=0.95\textwidth}]
\node {
\scriptsize
\centering \underline{Source}

\raggedright
Towns like Bluffton say they're working to promote communities of inclusiveness, but what does that really mean, and will people of color come? Jim Hunt is president of the National League of Cities based in Washington, D.C. He heads up the group's partnership for working toward inclusive communities. Mr. Hunt joins us from West Virginia Radio in Morgantown, West Virginia..And joining us via phone is Harvard law professor Charles Ogletree. Professor Ogletree leads the law schools' Charles Hamilton Houston Institute for Race and Justice. Gentlemen, great to have you on the program. Professor, let me start with you. Give us a quick historical view, of you will, of these, quote, sundown towns..Well, thanks, Ed. As you know, when I was doing research on the Tulsa race riots from 1921, I was astounded to learn of the literally hundreds of towns - now we've discovered thousands of towns around America - in the South, but not exclusively in the South, where blacks were told leave before sundown. There were sirens, there were notices, and the consequences of staying in those towns were death or other serious bodily injury..And the good news is that that is largely historical, but it's a frightening sense that people could not live in the community. They could work there, they could visit there, but they couldn't' be there after dark because of the strongly held feelings about segregation. And sundown towns were just that: if you're black, get out of town before the sun goes down..Mm hmm. Jim Hunt, while those sirens may have gone away, there are certainly still areas in this country where when the sun goes down, minorities know that this is an area - a region you should not be in. That continues today..Right, Ed. And I think when we look around - and, obviously, the inclusive agenda is a broader agenda than just race, but clearly race plays a very significant part in the program..Mr. Hunt, why are we seeing this sense of inclusiveness? Is it totally altruistic? Are there financial benefits to these cities? Why are we seeing this rush now to try to move to include minorities?.Well, and I think when we look across the board of what's happening across the country - when we look at Dr. Richard Florida's work on the rise of the creative class, and some of the economic development needs of cities throughout America - we recognize that if we're not going to be inclusive, we're going to suffer from an economic development perspective.

\centering \underline{Expert Speech Summary}

\raggedright
Some towns in America used to be known as sundown towns because black people were told to leave town before sundown or they could be harmed. Some cities today have programs that focus on being inclusive to minorities.

\centering \underline{Expert Text Summary}

\raggedright
Jim Hunt, President of the National League of Cities, and Harvard Law Professor Charles Olgetree discuss sundown towns and why some towns are working to be more inclusive of minorities. A sundown town was a town, mostly in the south, where blacks were allowed work and visit, but were not allowed to be there after sundown. If they were, the could face serious injury or even death. While these town have mostly gone away, there are still areas in the country that minorities know they should not be in when the sun goes down. Many towns are now working to be more inclusive of minorities and part of the reason has to do with the financial benefits they'll receive.

\centering \underline{Non-expert Speech Summary}

\raggedright
Jim Hunt, President of the National League of Cities, and Harvard law professor Charles Ogletree discuss the historical and contemporary implications of "sundown towns," where African Americans were forced to leave before sunset under threat of violence. While this practice is largely historical, there are still areas where minorities feel unsafe after dark. They emphasize the importance of inclusivity for economic development and societal well-being.

\centering \underline{Non-expert Text Summary}

\raggedright
The National League of Cities is working on inclusive communities, addressing historical issues like "sundown towns," where minorities were told to leave before dark under threats of violence. Harvard Law Professor Charles Ogletree discusses the historical prevalence of such towns, emphasizing the progress made. National League of Cities President Jim Hunt highlights the economic benefits of inclusivity in cities, linking it to overall economic development. While acknowledging past discrimination, the focus is on building inclusive communities for the future.

};
\end{tikzpicture}
\caption{\centering Example source and summaries.}
\end{figure*}

\section{Non-Expert annotation details}
\label{sec:appendix_nonexpert}

\subsection{IRB Review and Informed Consent}

Non-expert annotations were carried out using Amazon Mechanical Turk. The entire plan for this research paper was presented to an Institutional Review Board (IRB) and received assent before starting data collection. A consent form (shown next) was prepared and presented to prospective annotators, and conditional on their acceptance, they were allowed to take on the annotation task.

\includepdf[pages={1,2}]{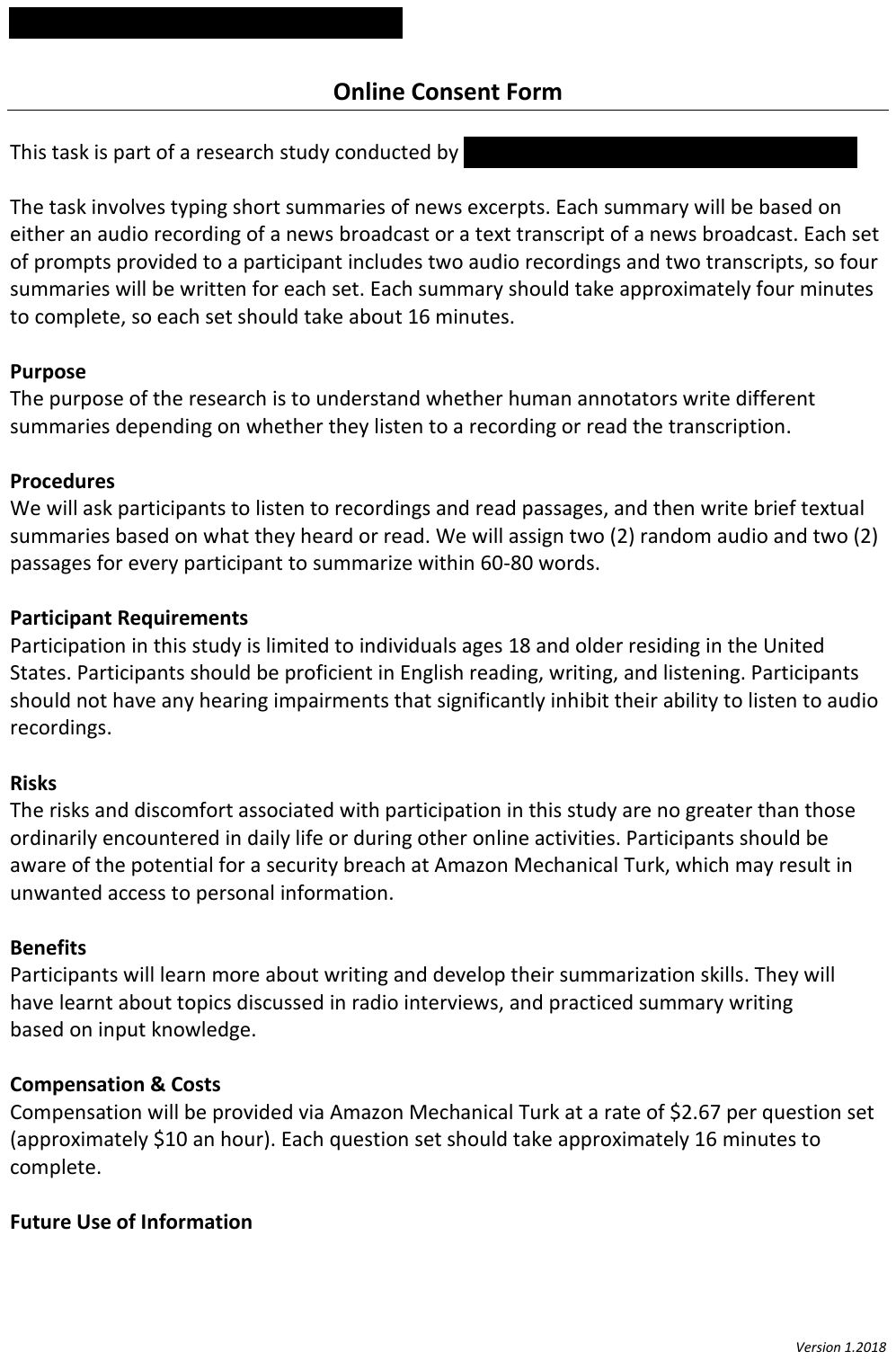}

\subsection{Compensation}

Workers were compensated at a rate of \$10/hour for their work, which is above the minimum wage in the United States of America, where the workers reside. 

\subsection{Worker Selection}
\begin{figure}
    \centering
    \includegraphics[scale=0.4]{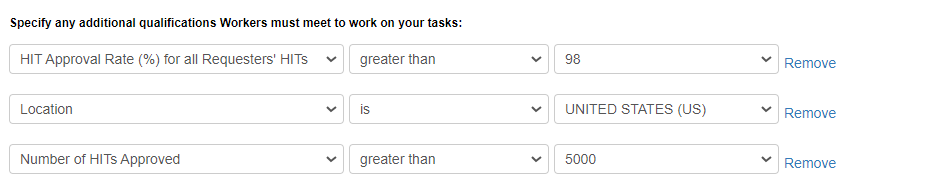}
    \caption{Filters to select workers.}
    \label{fig:workers-selection-criterion}
\end{figure}
Only the workers who agree to the consent form are allowed to perform the task on AMT. Furthermore, workers who reside in the US, with an AMT HIT approval rate of greater than 98\% and  total HITS approved greater than 5000 are selected for the task. Figure \ref{fig:workers-selection-criterion} shows the criterion used. 

\subsection{HIT Design and Guidelines}

\begin{figure}
    \centering
    \includegraphics[scale=0.4]{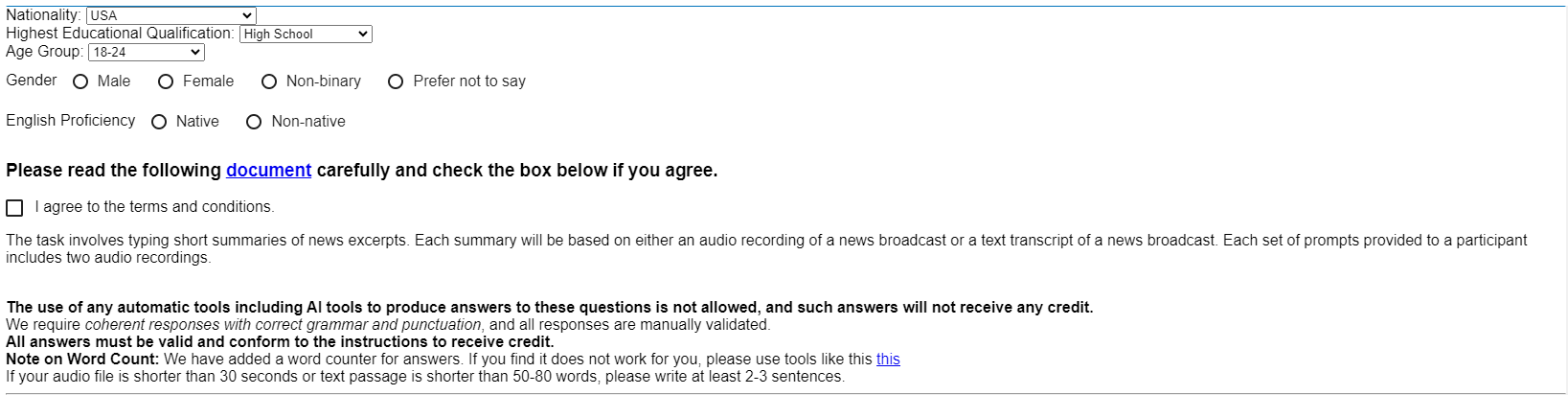}
    \includegraphics[scale=0.4]{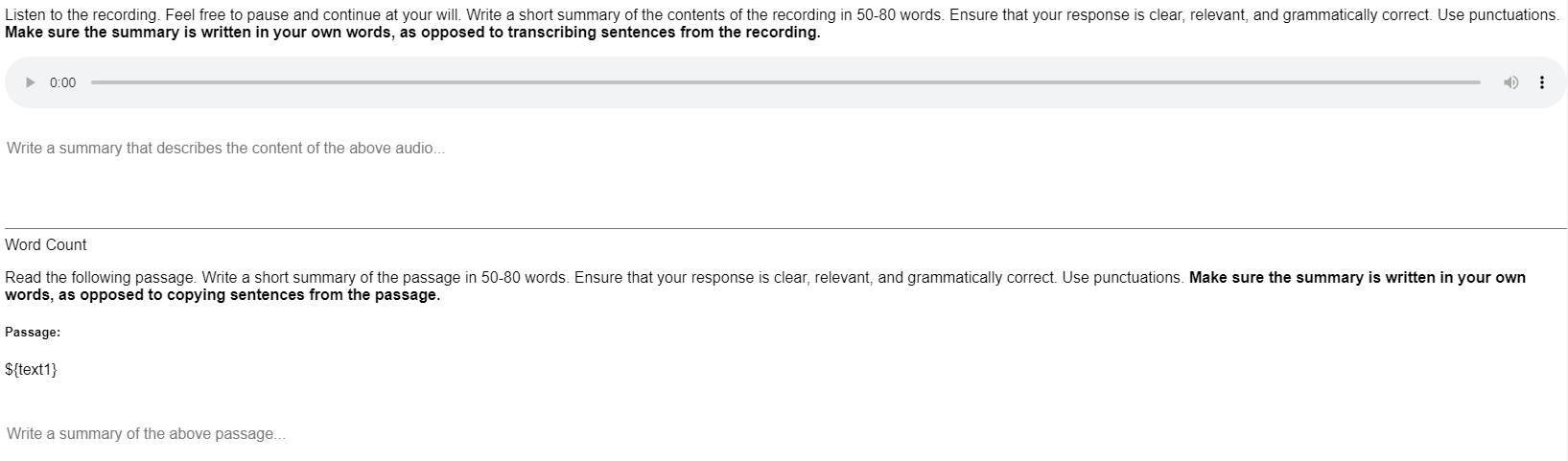}
    \caption{AMT design. Top figure shows the initial instructions presented to the workers. Bottom image shows how the audio and text source texts are presented.}
    \label{fig:amt_deisgn}
\end{figure}

Figure \ref{fig:amt_deisgn} shows the layout of the AMT design, how the instructions are presented, and how source audio and text are provided. The worker then proceeds to write the summaries in the text box below the source. 

As part of the instructions, the use of generative AI was prohibited to answer the questions.

Annotators were asked to write summaries based on the provided inputs while ensuring that the response was coherent and grammatical, and written in their own words.  All responses are validated manually for conformity to the instructions, and invalid responses are rejected and re-assigned to a new annotator. 

\subsection{Demographics of the AMT Workers}

\begin{figure}
    \centering
    \includegraphics[scale=0.4]{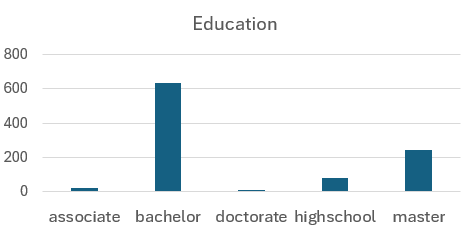}
    \includegraphics[scale=0.4]{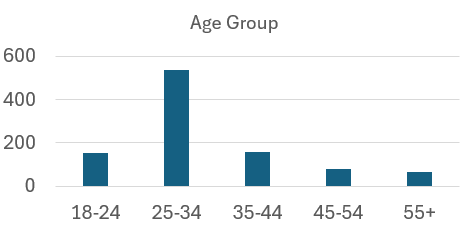}
    \includegraphics[scale=0.4]{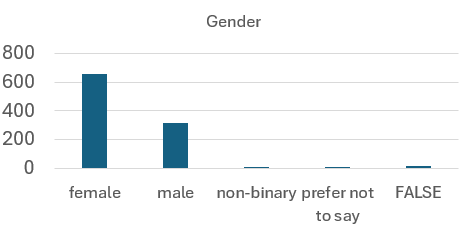}
    \caption{The demographics of the AMT workers are shown in the figure above. The variation in the education level, age groups and gender are shown as histograms.}
    \label{fig:amt_demographics}
\end{figure}

Figure \ref{fig:amt_demographics} shows the demographics of the AMT workers. The workers are mostly bachelor graduates, from the 25-34 age bracket, more female than other genders. 952 of the workers are native English speakers and 41 are not. 1000 of the workers are American nationals, whereas 3 are from elsewhere. We do not collect any information from the workers that can uniquely identify them. 

\subsection{Data Release}
We plan to release the summaries annotations collected from experts and non-experts for both modalities under the Apache 2.0 license, subject to the terms of the NPR license. Access to the data will be under the usage policy of NPR and the data is meant to be used for research purposes only.


\section{Results}




\subsection{Expert Annotations}

\begin{table*}[h]
    \small
    \centering
    \begin{tabular}{lccc}
    \toprule
          Metric& Transcript Overall& Speech Overall& p-value\\\toprule 
          Summary Length &  \textbf{78.58 $\pm$ 27.72}&  56.95 $\pm$ 18.30& \\
          Compression Ratio &3.76 $\pm$ 2.20&  \textbf{5.20 $\pm$ 3.06} & 8.83E-86\\
          Word Vocabulary Size &20,408&  16,211& -\\
          Novel Words \%&24.63	$\pm$ 13.57&  \textbf{25.66 $\pm$ 13.22} & 5.80E-02\\ 
          \midrule
          Extractiveness$\downarrow$ & 20.70 $\pm$ 10.53&  \textbf{18.57	$\pm$ 10.88}& 1.09E-08\\
          Entity F1$\uparrow$ &\textbf{33.63	$\pm$ 29.83}&  26.21	$\pm$ 26.90& 1.34E-30\\
          BERTScore $\uparrow$ & \textbf{85.66	$\pm$2.45} &  85.32	$\pm$2.42& 1.54E-05\\
          BARTScore $\uparrow$ & -3.37	$\pm$0.47&  -3.46 $\pm$ 0.45& 7.96E-11\\
          Retrieval Accuracy $\uparrow$ & \textbf{68.86 + $\pm$ 46.31}	&	59.04 $\pm$ 49.75 & 1.36E-19 \\
          MRR $\uparrow$ & 70.23 $\pm$ 16.08 & \textbf{75.74 $\pm$ 16.29} & 8.48E-52\\
          Factual Consistency (UniEval) $\uparrow$ & 0.83	$\pm$ 0.16&	\textbf{0.85$\pm$	0.16}&	1.90E-04\\ \midrule
          Pairwise Rep. (LLM) $\uparrow$ & \textbf{66.67} & 28.52 & -\\
          Pairwise Factualness (LLM) $\uparrow$ & \textbf{60.16} & 33.07 & -\\ 
          Pairwise Rep. (Human) $\uparrow$ & \textbf{48.57} & 21.43 &-\\ 
          Pairwise Factualness (Human) $\uparrow$ & \textbf{39.52} & 22.62 &-\\
          Fluency (UniEval) $\uparrow$ & \textbf{94.02$\pm$	4.54}	&94.01$\pm$	5.26&	9.83E-01\\
          Coherence (UniEval)$\uparrow$ &\textbf{91.82$\pm$	12.97}&	89.34$\pm$	15.63	&5.42E-08\\
          Pairwise BARTScore $\uparrow$ & -2.81$\pm$0.51&\textbf{-2.56$\pm$	0.52}&	1.67E-101\\ \bottomrule
    \end{tabular}
    \caption{RQ1 Evaluation : Speech versus Transcript-based Summaries by expert annotators}
    \label{tab:expert_rq1}
\end{table*}

\subsection{Non-Expert Annotations}
\label{sec:appendix_res_nonexp}

\begin{table*}[h]
    \small
    \centering
    \begin{tabular}{lccc}
    \toprule
            Metric&Text Overall&  Audio Overall& p-value\\\toprule 
            Summary Length & 75.35 $\pm$	17.70	& 77.42 $\pm$ 33.85	& 0.02\\
            Compression Ratio &3.69	$\pm$	1.95	& 3.66$\pm$		1.86	&0.78\\
            Word Vocabulary Size &21511	&	21676\\
            Novel Words \%& 37.47	$\pm$ 13.38&	38.45$\pm$	14.21&	0.09 \\ 
            \midrule
            Extractiveness (ROUGE-L with Transcript)& \textbf{18.37$\pm$	7.26}&	18.38	$\pm$11.29&	0.98 \\
            Entity F1& \textbf{34.51	$\pm$ 17.28}&	30.09$\pm$	17.72&	7.37E-14 \\
            BERTScore with Transcript & \textbf{85.28$\pm$	2.32} &	85.06 $\pm$2.76&	0.01 \\
            BARTScore with Transcript &\textbf{-3.41$\pm$	0.44}&	-3.43$\pm$	0.53&	0.30 \\
            Retrieval Accuracy& \textbf{71.82 $\pm$ 44.99} &  65.78 $\pm$ 47.44 & 1.04E-04 \\
            MRR & \textbf{76.56 $\pm$ 71.01} & 71.01 $\pm$ 41.57 & 5.80E-01 \\
            Factual Consistency (UniEval)& \textbf{82.08	$\pm$ 16.90}	& 78.57$\pm$	19.79	&1.49E-08\\
            \midrule
            Fluency (UniEval)& \textbf{92.90$\pm$	8.21} & 92.65	$\pm$9.35&	0.39\\
            Coherence (UniEval)& \textbf{90.88$\pm$	15.19}&	87.95$\pm$	19.57&	6.83E-07\\
            Pairwise BARTScore& \textbf{-2.85$\pm$	0.63 }&	-2.88$\pm$	0.70&	0.04\\ 
            \bottomrule
    \end{tabular}
    \caption{Non-Expert Annotation}
    \label{tab:non_expert_annotations}
\end{table*}

\end{document}